\documentclass[a4paper,fleqn]{cas-dc}
\usepackage[numbers,sort&compress]{natbib}
\usepackage{booktabs,amsmath,amssymb}

\makeatletter
\let\scvibOriginalFirstPageNotes\printFirstPageNotes
\renewcommand{\printFirstPageNotes}{%
  \begingroup
  \let\@footnotetext\H@@footnotetext
  \scvibOriginalFirstPageNotes
  \endgroup
}
\makeatother

\ExplSyntaxOn
\RenewDocumentEnvironment { Abstract } { o }
{
  \group_begin:
  \IfNoValueTF { #1 } { } { \tex_gdef:D \abstractname { #1 } }
  \parindent 0pt
  \box_if_empty:NTF \g_stm_key_box
    { \leftskip = .35 \textwidth }
    {
      \dim_gset:Nn \l_tmpa_dim { \box_ht:N \g_stm_key_box }
      \dim_gadd:Nn \l_tmpa_dim { \box_dp:N \g_stm_key_box }
      \leftskip .35\textwidth
      \hspace*{-.35 \textwidth}
      \noindent\hbox_to_wd:nn { 0pt } { \box \g_stm_key_box \hss }
      \skip_vertical:n { - \l_tmpa_dim }
    }
  \noindent \abstractname \par
  \skip_vertical:n { -4pt }
  \noindent \rule{.65\textwidth}{.2pt}\par \footnotesize
  \ignorespaces \everypar { \parindent=1.5em }
}
{ \par \group_end: }
\ExplSyntaxOff
\usepackage{placeins}
\renewcommand{\dbltopfraction}{0.78}

\usepackage{array,tabularx,siunitx}
\newcommand{\paperTableStyle}{\renewcommand{\arraystretch}{1.18}\setlength{\tabcolsep}{3pt}}
\ExplSyntaxOn
\cs_set:Npn \__first_footerline:
{
  \group_begin:
  \small
  \sffamily
  \ifnum\theblind>0\relax\else\__short_authors: :~\fi
  {\rmfamily\itshape Preprint}
  \group_end:
}
\ExplSyntaxOff
\begin{document}
\let\printorcid\relax
\shorttitle{Editable state-conditioned visual instance binding}
\shortauthors{Yang et al.}
\title[mode=title]{SCVIB: Editable State-Conditioned Visual Instance Binding for Multi-Turn Personalized Localization}
\author[1]{Xiongtai Yang}\ead{2025223045170@stu.scu.edu.cn}
\author[1]{Ziyan He}\ead{2025223045235@scu.edu.cn}
\author[1]{Tao Wang}\cormark[1]\ead{twangnh@gmail.com}
\affiliation[1]{organization={Sichuan University},addressline={No. 24, South Section 1, First Ring Road, Wangjiang Campus},city={Chengdu},postcode={610065},state={Sichuan},country={China}}
\cortext[1]{Corresponding author}
\begin{abstract}
Stored visual evidence can remain unchanged while its relevance to a current request changes with interaction state. To study this state-dependent source relevance, we introduce SCVIB, a controlled diagnostic benchmark for editable state-conditioned visual instance binding. SCVIB contains 1,050 manually verified support--query pairs and 1,500 multi-turn episodes across five visual domains and four target-state dependencies. Executable state semantics separate target resolution from evidence selection and query-side localization. Direct Qwen3-VL-8B inference achieves 99.53\% target-resolution accuracy but 62.00\% Joint@0.5, revealing a 37.53-point resolution--localization gap; 563 of its 570 joint failures follow correct target identification. Routing comparisons and fixed-checkpoint support interventions diagnose a post-resolution evidence-utilization bottleneck: successful resolution must be coupled with relevant visual evidence and effective query-side grounding. We introduce TT-VG (Transition-Trace Visual Grounding), a state-conditioned visual grounding framework with an explicit and auditable state-to-evidence-to-grounding path. Target-State Transition Trace (TSTT) executes editable state and selects an auditable support record; Visual Evidence Grounding Adaptation (VEGA) learns support-conditioned grounding from target appearance, boxed support context, and the query. The canonical TT-VG configuration reaches 69.80\% Joint@0.5, improving over direct inference by 7.80 points at the combined evidence-utilization and grounding interface. Task adaptation improves the DINOv2 and MASA retrieval baselines to 51.47\% and 55.67\%, respectively, but both remain substantially below VEGA. These findings highlight state-dependent source selection and evidence utilization as consequential interfaces for interactive multimodal information fusion.
\end{abstract}
\maketitle
\begin{figure}[pos=!t]
\centering
\includegraphics[width=\linewidth]{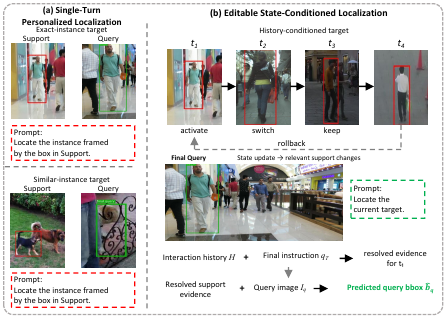}
\caption{Fixed-target versus editable state-conditioned localization. Left, a support specifies one target; right, state edits and rollback select relevant support. Red and green boxes mark support annotations and query targets.}
\label{fig:teaser}
\end{figure}

\section{Introduction}

In a continuing interaction with a visual assistant, a user may change or restore the current target without restating its full visual specification at every query. Consider stored support images of a backpack, a suitcase, and a bicycle: the user may focus on the backpack, switch to the suitcase as the search changes, and undo that switch to resume the earlier target. Editable interaction state maintains this current-target context while the support images remain available, allowing later requests to refer to the current target without re-uploading evidence. The stored evidence remains unchanged, but the support relevant to localization changes with the user's edits. This state-dependent source relevance makes interactive localization a problem of information selection and visual grounding (Figure~\ref{fig:teaser}).

Visual grounding and open-vocabulary detection connect category labels or referring expressions to image regions~\citep{Kamath2021MDETR,Li2022GLIP}; personalized localization uses an annotated support example to specify a concrete instance~\citep{Doveh2025IPLoc}. In multi-turn interaction, end-to-end multimodal inference must resolve the current target, select its support evidence, utilize that evidence, and localize the same instance in a new scene. A single prediction leaves these intermediate decisions implicit. When localization fails, the output alone does not identify where this chain broke.

\begin{figure*}[pos=!t]
    \centering
    \includegraphics[width=\textwidth]{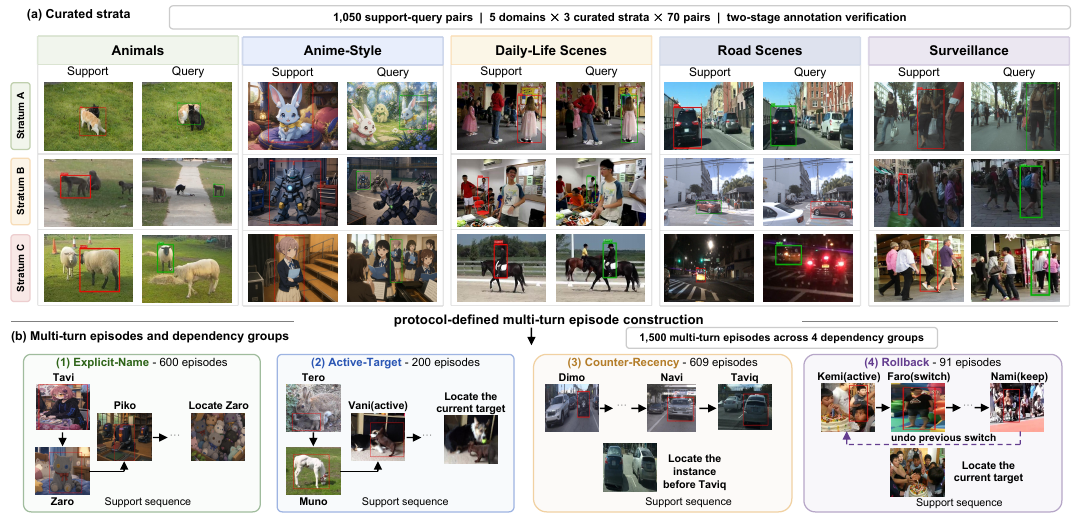}
    \caption{SCVIB construction and target-state dependencies. Upper: 1,050 pairs across five domains and three curated strata. Lower: four dependencies, with red support boxes and illustrative green query targets. The pairs yield 1,500 controlled episodes.}
    \label{fig:overview}
\end{figure*}

We introduce SCVIB as a controlled diagnostic benchmark that makes this chain measurable. Its 1,050 manually verified support--query pairs generate 1,500 episodes across five visual domains and three curated strata. Episode-local pseudo-names bind instances to stored evidence, and visible events define executable state transitions. The four dependencies are Explicit-Name, Active-Target, Counter-Recency, and Rollback. They cover direct lookup, active-state maintenance, non-latest source selection, and restoration (Figure~\ref{fig:overview}). The closed interaction grammar provides experimental control: each episode has a unique executable target reference, enabling failure attribution across target resolution, evidence selection, and query-side localization within each controlled episode.

This decomposition reveals a resolution--localization gap. Direct Qwen3-VL-8B inference reaches 99.53\% target-resolution accuracy but 62.00\% Joint@0.5, a 37.53-point difference. Of its 570 joint failures, 563 occur after correct target identification. Routing the predicted identity to a support-conditioned grounder raises Joint@0.5 to 69.33\%; fixed-checkpoint support interventions further show that changing the supplied evidence can sharply reduce localization. Together, these observations diagnose a post-resolution evidence-utilization bottleneck: correct resolution must be connected to relevant visual evidence and effective grounding of the requested instance in the query image.

We introduce TT-VG (Transition-Trace Visual Grounding), a state-conditioned visual grounding framework that connects resolved interaction state to relevant evidence and downstream localization through explicit state-to-evidence execution and support-conditioned grounding. Target-State Transition Trace (TSTT) executes visible state edits and resolves the final reference to an auditable support record. A Visual Evidence Package organizes selected visual evidence into target-focused appearance, boxed support context, and the query scene. Visual Evidence Grounding Adaptation (VEGA) learns support-conditioned grounding from this package. The canonical TT-VG configuration reaches 69.80\% Joint@0.5, improving over direct inference by 7.80 points at the combined evidence-utilization and grounding interface. Task-adapted DINOv2 and MASA controls test whether access to task training explains the localization advantage, while dependency analysis identifies the dependencies that benefit from explicit state restoration.

In this interactive setting, a central question for information fusion is which source to integrate as interaction state changes. Language determines the current request; stored supports provide instance-specific appearance and context; the query supplies the scene for localization. SCVIB controls the relevance decision and measures its downstream consequences. TT-VG executes this selection explicitly before integrating the selected visual inputs.

We make three contributions. (1) We introduce the controlled diagnostic benchmark SCVIB with executable state semantics and separate target-resolution and localization measurements. (2) We identify a resolution--localization gap and diagnose a post-resolution evidence-utilization bottleneck through failure decomposition, routing comparisons, support interventions, and selected-backend adaptation. (3) We introduce TT-VG, a state-conditioned visual grounding framework that explicitly connects resolved interaction state to relevant evidence and downstream localization, combining executable state semantics, selected support records, and learned query-side grounding of the selected instance.

\section{Related Work}
\subsection{Visual Grounding and Personalized Localization}
Visual grounding and open-vocabulary detection connect language to image regions. CLIP learns transferable image--text representations~\citep{Radford2021CLIP}; MDETR, GLIP, RegionCLIP, and Grounding DINO develop region-level alignment and localization~\citep{Kamath2021MDETR,Li2022GLIP,Zhong2022RegionCLIP,Liu2025GroundingDINO}. Recent Information Fusion reviews organize advances in 2D/3D grounding and multimodal large-model object detection~\citep{Guo2026VGSurvey,Sapkota2026Detection}. These directions provide the cross-modal representations and localization tools underlying support-conditioned instance matching.

Personalized localization uses annotated visual support to specify the target instance. IPLoc adapts VLMs on same-instance pairs derived from trajectories~\citep{Doveh2025IPLoc}. FOCUS uses support-region attention supervision and policy optimization for category-agnostic in-context localization~\citep{Karim2026FOCUS}. These methods study localization conditioned on a given support target; SCVIB additionally controls how editable interaction state determines which stored support is currently relevant. POLAR personalizes representations through regularized low-rank updates~\citep{Ryan_2025_CVPR}. PerSAM uses a support reference for segmentation, MASA matches instances across images, and PDM studies diffusion features for personalized segmentation and retrieval~\citep{Zhang2024PerSAM,Li2024MASA,Samuel2024PDM}. Identity persistence in re-identification and tracking offers related supervision, exemplified by Market-1501, LaSOT, and TAO~\citep{Zheng2015Market1501,Fan2019LaSOT,Dave2020TAO}. SCVIB builds on support-defined targets and adds editable state-dependent source selection: several episode-local bindings coexist, and visible state updates govern which evidence is relevant to final localization.

\subsection{Selective Multimodal Information Fusion}
Multimodal fusion studies how heterogeneous sources are aligned, represented, and integrated~\citep{Lahat2015Fusion,Baltrusaitis2019Survey}. Cross-modal attention, as in MulT, links information across unaligned streams~\citep{Tsai2019MulT}. Dynamic fusion can weight modalities according to input quality and uncertainty~\citep{Zhang2023QMF}; a recent Information Fusion survey organizes noisy, incomplete, imbalanced, and quality-varying multimodal inputs~\citep{Zhang2026LowQualityFusion}. These perspectives motivate selective integration, with source contributions depending on their conditions of use.

Within visual grounding, U2ENet combines semantic and visual features for monocular 3D grounding~\citep{Guo2026U2ENet}, while modality-specific grounding offers a complementary approach to monocular 3D grounding~\citep{Qiu2026Grounding}. SCVIB studies state-dependent source relevance among stored support records. The stored support images can remain unchanged while a visible interaction edit changes which record should be integrated. TT-VG makes this selection executable and auditable, then uses VEGA to integrate the selected appearance, support context, and query scene. This formulation connects selective information fusion to personalized localization through a measurable state-to-evidence interface.

\subsection{Multi-Image and Long-Context MLLMs}
Flamingo supports interleaved visual and textual context for few-shot learning~\citep{Alayrac2022Flamingo}, and MANTIS develops multi-image skills through instruction tuning~\citep{Jiang2024Mantis}. Multi-image evaluation examines several forms of information use: MC-Bench studies multi-context grounding~\citep{Xu2025MCBench}; Mementos evaluates reasoning over image sequences; ReMI evaluates reasoning with multiple images; and MM-NIAH and MileBench probe multimodal context recall and long-context understanding~\citep{Wang2024Mementos,Kazemi2024ReMI,Wang2024MMNIAH,Song2024MileBench}. SCVIB connects these concerns to an instance-level output: an editable history determines the requested support record, and success requires localizing its corresponding instance with a query-side box. Separate target-resolution and joint-localization metrics reveal the resolution--localization gap and help analyze how visual evidence is used for localization after the requested instance has been identified.

\subsection{Interactive Multimodal Systems}
Visual Dialog studies answering questions using an image and dialogue history~\citep{Das2017VisualDialog}; GuessWhat?! identifies an object through a sequence of questions~\citep{DeVries2017GuessWhat}. Guessing State Tracking represents evolving beliefs over candidate objects during dialogue~\citep{Pang2020StateTracking}. Referential MLLMs such as Shikra and Ferret connect conversational references to spatial coordinates and regions~\citep{Chen2023Shikra,You2024Ferret}. Temporal grounding also exploits structured language cues, including action-guided prompt tuning for video grounding~\citep{Wang2025Action}. SCVIB focuses on explicit edits to episode-local target bindings across support images. Its accepted state events define an executable target, including switching and rollback, and its final output requires same-instance localization in a separate query image. This operational specification supports auditing the path from a visible edit to the evidence supplied to the grounder.
\section{SCVIB: Task and Benchmark}

\subsection{Problem Formulation}

Single-turn personalized localization takes a support image, a query
image, and a support-side bounding box---rendered as a red box---that
identifies the target instance. We extend this setting to editable
state-conditioned visual instance binding over multiple turns and
instantiate it as the SCVIB benchmark. The $i$-th support turn
introduces an instance tuple $(I_i^s, b_i^s, n_i)$, where $n_i$ is an
episode-local pseudo-name. Protocol-defined events---bind, activate,
switch, keep, discard, and rollback---govern evidence registration and
the evolution of target state $s_t$. In particular, bind registers a
support-defined instance without changing the active target.

The final query may refer to an explicit pseudo-name, the currently
active target, or an instance specified relative to the support order.
The model resolves the reference from the interaction history and returns a normalized query-image box for the requested target instance.

SCVIB is a controlled diagnostic benchmark: episodes are generated under
a closed protocol grammar and rendered as natural-language interactions.
The grammar controls state semantics and reference ambiguity. Visible
updates map deterministically to protocol events, yielding a unique state trace per episode for stage-specific failure analysis.

\subsection{Benchmark Construction and Episode Design}
SCVIB contains 1,050 support--query base pairs spanning five
visual domains and three model-assisted curated strata,
with 70 pairs per domain--stratum combination. During construction,
Qwen3-VL-4B single-turn localization supplied a model-assisted selection signal
\citep{Bai2025Qwen3VL}, with intended thresholds of IoU $\geq0.5$,
$0.1\leq\mathrm{IoU}<0.5$, and IoU $<0.1$. The labels Strata A, B, and C identify these three model-assisted
curated groups in the same order; their assignment reflects the
construction model and prompt.
Appendix~\ref{app:historical-audit} documents the fixed strata and their construction, together with the calibration results for individual benchmark pairs.

The benchmark combines tracking, driving, daily-scene, and surveillance imagery with 210 manually annotated Anime-Style pairs generated with OpenAI's GPT Image 2 API. Appendix~\ref{app:source-access} specifies source counts, acquisition procedures, and reconstruction recipes.

We normalize all pairs to a common
support--query localization format and construct 1,500 multi-turn
episodes with controls over visual domain, curated stratum, number of
supports, target position, and distractor mode. The resulting episode
set comprises 600 Explicit-Name, 200 Active-Target, 609
Counter-Recency, and 91 Rollback episodes, requiring explicit pseudo-name lookup, active-state maintenance, non-latest support selection, and restoration of an earlier target state, respectively.

Visual sampling is controlled independently of episode construction.
The evaluation set contains 120 episodes with two supports, 780 with
three, and 600 with five. Across episodes, the target appears at the first, an interior, or the
last support position in 581, 472, and 447 cases, respectively.
Pseudo-names are drawn from an 88-name inventory and assigned
independently across episodes, encoding no category, position, or
identity semantics. Each base pair contributes to at most two episodes.

\subsection{Data Verification and Evaluation Protocol}

We use a two-stage task-validity and annotation curation process. For the
four real-image domains, one curator screens candidate pairs
for instance consistency, target visibility, ambiguity, and
bounding-box quality, and a second verifies all retained pairs.
Anime-Style pairs are manually boxed before the same second-stage verification.

Beyond annotation quality, we control hidden-answer leakage during routing and source overlap between training and evaluation. No
evidence-selection mechanism receives query-side ground-truth boxes,
hidden target annotations, target-turn metadata, or ground-truth
support indices. Re-executing the deterministic resolver from the
visible interaction history and final query reproduces the same
evidence record for all 1,500 episodes. The LaSOT-derived training and evaluation pairs use disjoint source videos.

Evaluation uses mIoU, BBox Acc@$\tau$, Target Resolution Accuracy (TR), and
Joint@$\tau$. Over all episodes, mIoU is the mean query-side box IoU,
whereas BBox Acc@$\tau$ is the proportion of predictions with IoU
$\geq \tau$. Both localization metrics evaluate all predictions regardless of whether the target identity is correct.

Target Resolution Accuracy measures whether a method identifies the
ground-truth support-defined instance. For Seq-free and Structured
State-Table Prompting, correctness is based on the predicted
episode-local pseudo-name, with invalid or incorrect names counted as
errors. For configurations with an explicit evidence-selection
mechanism, it is based on the selected evidence record. Because each
pseudo-name maps uniquely to one evidence record within an episode,
both interfaces are evaluated against the same support-defined
instance. Target Resolution Accuracy measures final-reference resolution.
Joint@$\tau$ is the primary end-to-end metric, requiring both correct
target resolution and query-side box IoU $\geq \tau$.

We call the difference between Target Resolution Accuracy and
Joint@0.5 the resolution--localization gap, measured in percentage
points. Joint success is a subset of correct target resolutions, so
this descriptive difference counts episodes with correct target
identification and unsuccessful query-side localization. Invalid boxes receive zero IoU and count as localization failures.

\FloatBarrier
\section{TT-VG: State-Conditioned Visual Grounding Framework}
\begin{figure*}[pos=tb]
  \centering
  \includegraphics[width=\textwidth]
  {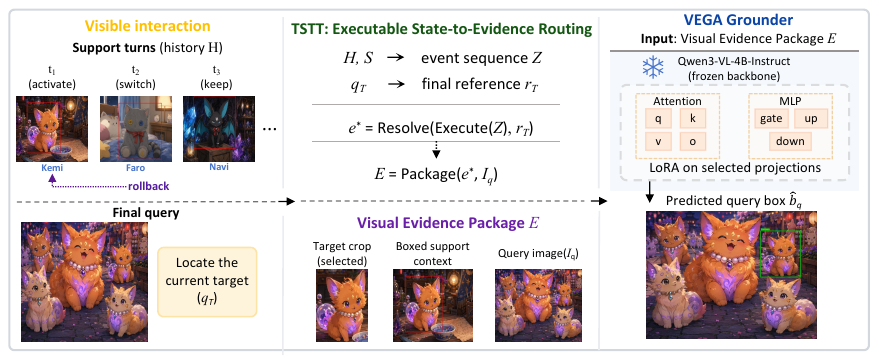}
  \caption{TT-VG makes the state-to-evidence-to-grounding path explicit. TSTT selects the evidence record, the Visual Evidence Package combines its crop and boxed context with the query, and VEGA predicts the query box. The selected evidence is inspectable before localization.}
  \label{fig:method}
\end{figure*}

TT-VG connects resolved interaction state to query-side localization through two explicit
interfaces: editable state to selected evidence, and selected evidence
to query-side localization. TSTT executes the visible interaction and
returns a traceable support record. VEGA learns support-conditioned
grounding from that record's Visual Evidence Package. The framework
makes the path from a requested identity to its visual evidence inspectable before localization (Figure~\ref{fig:method}).

TSTT implements the protocol-defined state semantics with an evidence registry,
versioned active states, and a switch-undo stack. It compiles visible
updates into protocol events and resolves the final-query reference
against the executed state. Let $H$ denote the visible interaction
history before the final query, $q_T$ the visible final-query
instruction, $S=\{(I_i^s,b_i^s,n_i)\}$ the support set, and $I_q$ the
query image. TSTT selects the evidence record, and VEGA predicts the
corresponding query box. Figure~\ref{fig:method} shows this path.

\subsection{TSTT: Target-State Transition Trace}
Within TT-VG, TSTT implements an executable and auditable mapping from interaction state to selected evidence. It connects the visible interaction history and final reference to the support record used for grounding.

\noindent\textbf{Event compilation.}
The visible interaction history $H$ and support set $S$ yield a protocol-defined event sequence:
\begin{equation}
Z = \mathrm{CompileEvents}(H,S).
\label{eq:event-compilation}
\end{equation}
The sequence $Z$ records the support-introduction and state-update
events expressed in the visible interaction.

\noindent\textbf{Reference compilation.}
The final-query instruction is compiled separately into the
final-query reference:
\begin{equation}
r_T = \mathrm{CompileReference}(q_T).
\label{eq:reference-compilation}
\end{equation}
The reference $r_T$ retains only information expressed in $q_T$ and is
resolved only after $Z$ has been executed. This separation prevents
$r_T$ from altering event execution.

\noindent\textbf{Versioned state execution.}
TSTT executes the event sequence over an evidence registry and versioned states:
\begin{equation}
\mathcal{T} = \mathrm{Execute}(Z),
\label{eq:state-execution}
\end{equation}
where $\mathcal{T}$ denotes the executed Target-State Transition Trace.

The registry stores one evidence record per support turn. Each record
associates an episode-local pseudo-name with its support image and
support-side bounding box. Each versioned state record tracks the
active-evidence pointer and the predecessor information required by the
interaction protocol. At each support turn, TSTT first registers the evidence, then applies the corresponding protocol-defined state update.

The bind and keep events preserve the active-evidence pointer, whereas
activate and switch update it. The discard event registers the introduced
instance as a distractor without changing the current target. Rollback
follows the stored history associated with the latest switch and restores
the preceding active-evidence pointer. Rollback leaves every registered evidence record intact.

\noindent\textbf{Final-query reference and evidence resolution.}
After event execution, TSTT resolves the final-query reference against
the resulting target state and evidence registry:
\begin{equation}
e^\star = \mathrm{Resolve}(\mathcal{T}, r_T),
\label{eq:evidence-resolution}
\end{equation}
where $e^\star$ is the resolved evidence record. It links the final reference to the support selected from the executed state.

The resolution rule follows the reference form encoded in $r_T$. An
explicit pseudo-name triggers direct registry lookup, an active-target
reference follows the final active-evidence pointer, and a relative
reference is evaluated over the registered support order. The emitted record is passed to support packaging and then to VEGA to localize the selected instance in the query image.

\subsubsection{State Operations and Execution Semantics}
TSTT uses six canonical event types. Bind maps to
\texttt{introduce\_\allowbreak{}only}; activate maps to \texttt{set\_\allowbreak{}active} at
introduction or \texttt{mark\_\allowbreak{}active} for an already registered name;
switch maps to \texttt{switch\_\allowbreak{}active}; keep and distractor/discard
updates map to \texttt{keep\_\allowbreak{}active} or \texttt{introduce\_\allowbreak{}only},
depending on the rendered template; rollback maps to
\texttt{undo\_\allowbreak{}previous\_\allowbreak{}switch}. ``Discard'' preserves a distractor
in the evidence registry while excluding it from the active pointer.

For example, introducing A as active, switching to B, and then introducing
C as a distractor leaves B active. Undo restores A as active while A, B,
and C remain registered. An explicit final query for C is consequently
still meaningful. The state versions retain predecessor links and the
switch rollback stack, forming an execution trace from visible updates
to the final selected support record.

The main benchmark uses a closed rendering grammar with deterministic resolution for accepted commands. Unsupported updates,
activation of a not-yet-introduced name, duplicate names, contradictory
switch origins, and undo without a prior switch are rejected. A language
parser can provide an additional front end, but its errors must be evaluated
separately from deterministic state execution.

\subsection{VEGA: Visual Evidence Grounding Adaptation}

VEGA learns support-conditioned instance grounding from a structured Visual Evidence Package consisting of target-focused appearance, boxed support context, and the query scene. The package supplies the selected instance and its scene context to TT-VG's learned grounder. Each crop emphasizes local instance appearance, whereas the boxed support context preserves spatial and contextual cues. The pseudo-name serves only as an episode-local identifier; VEGA is conditioned on the resolved support evidence.

The resolved evidence record and query image form the Visual Evidence Package $E$ used for grounding:
\begin{equation}
\begin{array}{rcl}
E &=& \mathrm{Package}(e^\star, I_q),\\
\hat{b}_q &=& \mathrm{VEGA}(E),
\end{array}
\label{eq:vega-inference}
\end{equation}
where $\hat{b}_q$ is the predicted query-side bounding box.

The resolved evidence record $e^\star$ associates an
episode-local pseudo-name with its support image and
support-side bounding box. The packaging layer derives a
target crop and boxed support context from this record. The
Visual Evidence Package combines these local and contextual
support views with the query image and the localization and
output-format instruction. The canonical multi-turn package uses a tight crop. Training uses 3\% padding around the clipped support box; the appendix specifies each diagnostic's rendering.

VEGA uses Low-Rank Adaptation (LoRA) to adapt Qwen3-VL-4B-Instruct \citep{Bai2025Qwen3VL,Hu2022LoRA}. We construct a fixed training
set of 5,000 trajectory-derived same-instance support--query pairs
from LaSOT \citep{Fan2019LaSOT}. In each pair, a support annotation
identifies a concrete instance to be localized in the query image
across changes in object scale, viewpoint, appearance, and surrounding visual context.

For the $j$-th training pair $(I_j^s,b_j^s,I_j^q,b_j^q)$, the target crop is $C_j^s=\mathrm{Crop}(I_j^s,b_j^s)$ and the boxed context is $\widetilde{I}_j^s=\mathrm{Box}(I_j^s,b_j^s)$. The training input and target are
\begin{equation}
\begin{array}{rcl}
V_j &=& [C_j^s,\widetilde{I}_j^s,I_j^q],\\
X_j &=& \mathrm{Format}(V_j,p_{\mathrm{loc}}),\\
Y_j &=& \mathrm{Serialize}\!\left(
\mathrm{Norm}_{1000}(b_j^q)\right).
\end{array}
\label{eq:vega-training-io}
\end{equation}
The visual inputs follow the order target crop, boxed support context,
and query image; $p_{\mathrm{loc}}$ denotes the fixed localization and
output-format instruction. The target $Y_j$ is the normalized
query-side bounding box serialized in the required format.
The visual roles and ordering match inference; training uses the padded crop convention specified above.

The final checkpoint is fixed for Random-Support + VEGA, Latest-Support + VEGA, Pred-Name Router + VEGA, and TT-VG. These controls use identical packaging and differ only in the selected evidence record.

\subsubsection{Training and Supervision}
We optimize the LoRA adapter for one epoch on the fixed 5,000-pair LaSOT training set, using answer-token cross-entropy on the serialized query box. Prompt and padding tokens are masked from the loss, so supervision covers only the serialized coordinate answer for the supplied support--query pair. Appendix~\ref{app:vega-training} details optimization, rendering, and the random seed.

\FloatBarrier
\section{Experiments}
\subsection{Experimental Setup}
We evaluate the full 1,500-episode benchmark using the canonical visible interactions and final-query references. Seq-free Qwen3-VL-8B-Instruct receives the rendered multi-image dialogue and predicts the final pseudo-name and query box. Pred-Name Router + VEGA maps that predicted name to a support record and passes the resulting Visual Evidence Package to the fixed VEGA checkpoint. TT-VG uses TSTT to resolve the support record and the same VEGA checkpoint for grounding. These three configurations separate direct full-history inference, predicted-name routing, and executable state resolution.

The localization comparison fixes target resolution for DINOv2 retrieval, MASA-R50 retrieval, box-initialized PerSAM, and VEGA. Each method receives the same TSTT-resolved support record and query. Grounding DINO proposals are ranked by DINOv2 or MASA features; PerSAM converts a support-conditioned mask to a query box~\citep{Liu2025GroundingDINO,Oquab2023DINOv2,Li2024MASA,Zhang2024PerSAM}. VEGA is adapted on 5,000 trajectory-derived LaSOT pairs; the external backends in the main localization table and primary direct MLLM are frozen. Retrieval-component adaptation is evaluated in Section~\ref{sec:backend-adaptation} and Table~\ref{tab:backend-adaptation}. Additional state-table, support-selection, and frozen 4B configurations appear in Appendix~\ref{app:extended-controls}.

All configurations are evaluated against the same continuous query annotations. Names identify support records; boxes use query-relative coordinates normalized to 0--1000. Failed generations and invalid names or boxes remain in the denominator. Direct final generation is greedy with a 128-token cap; VEGA is greedy with a 64-token cap. We report Joint@0.5 as the primary metric, together with Joint@0.7, mIoU, and Target Resolution Accuracy. Paired 95\% intervals use 10,000 episode resamples and 5,000 target-base-pair cluster resamples, preserving episodes that share a target pair. Appendix~\ref{app:reproducibility} provides full prompts, parsing and backend settings, and sensitivity analysis for shared query images.

\subsection{Main Results and Diagnostic Analysis}
We organize the main analysis around four questions: whether correct target resolution yields successful localization, how selected evidence affects localization, whether task adaptation explains the grounding advantage, and how explicit execution supports state restoration.
\par
\begin{table}[pos=htbp]
\caption{Query-side localization on 1,500 episodes at 100.00\% target resolution under stated training conditions. All methods receive the same selected support and query. Scores are percentages; bold denotes the best. Table~\ref{tab:backend-adaptation} provides retrieval adaptation controls using the same training pairs.}\label{tab:localization-comparison}
\centering\small\paperTableStyle
\begin{tabular*}{\linewidth}{@{\extracolsep{\fill}}lS[table-format=2.2]S[table-format=2.2]S[table-format=2.2]@{}}\toprule
Method & {Joint@0.5} & {Joint@0.7} & {mIoU} \\
\midrule
GDINO--DINOv2 & 49.33 & 41.53 & 48.34 \\
GDINO--MASA-R50 & 53.80 & 48.60 & 50.07 \\
Box-initialized PerSAM & 31.20 & 28.53 & 31.40 \\
VEGA & \bfseries 69.80 & \bfseries 62.33 & \bfseries 61.52 \\
\bottomrule\end{tabular*}
\end{table}

\subsubsection{Resolution--Localization Gap}
\begin{figure*}[pos=t]
\centering\includegraphics[width=\textwidth]{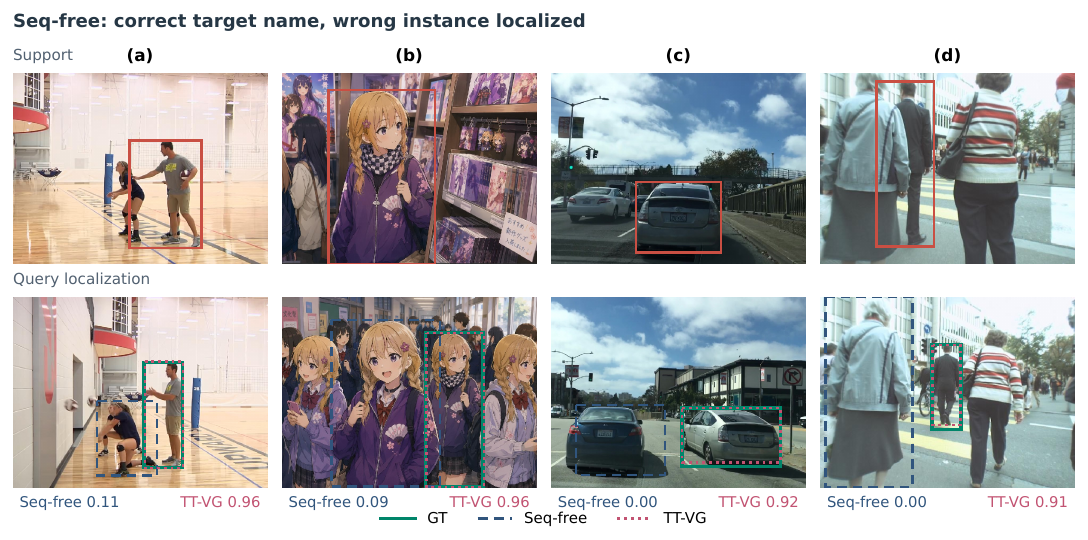}
\caption{Qualitative examples of localization errors after correct target identification. Top: support reference (red annotation); bottom: query ground truth (solid green), Seq-free (dashed navy), and TT-VG (dotted rose). Supports are explanatory for Seq-free; display crops retain all boxes, and IoUs use full-image coordinates.}\label{fig:seqfree-failures}
\end{figure*}

\begin{figure*}[pos=t]
\centering\includegraphics[width=\textwidth]{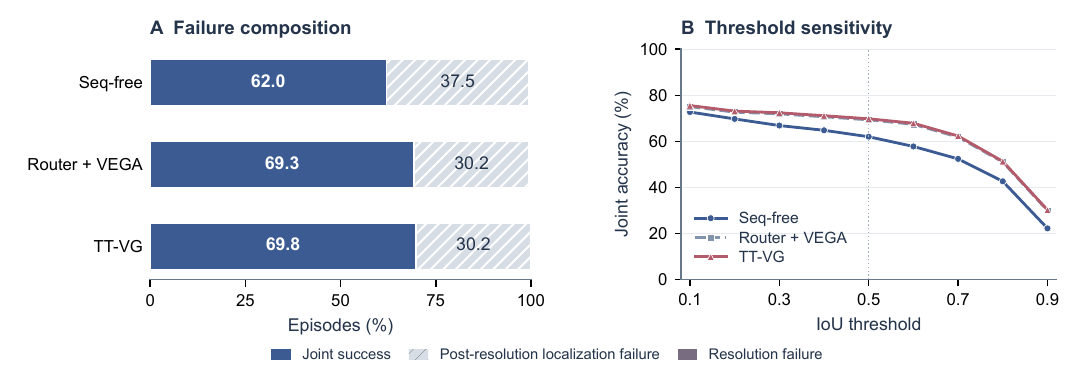}
\caption{Failure composition and threshold sensitivity on 1,500 episodes. Stacked bars separate joint success, post-resolution localization failure, and resolution failure; curves retain all outputs across IoU thresholds. Most direct-inference failures occur after correct target identification.}\label{fig:grounding-diagnostics}
\end{figure*}

Table~\ref{tab:main-results} reveals the resolution--localization gap: Seq-free reaches 99.53\% Target Resolution Accuracy and 62.00\% Joint@0.5, a difference of 37.53 points. Of 570 joint failures, 563 follow correct target identification and seven fail target resolution (Figure~\ref{fig:grounding-diagnostics}). Figure~\ref{fig:seqfree-failures} gives qualitative examples of Seq-free localizing the wrong instance after correctly identifying the target name.
\par
\begin{table}[pos=htbp]
\caption{Three-stage results on 1,500 episodes. Percentages are reported; Router + VEGA and TT-VG share VEGA weights.}\label{tab:main-results}\label{tab:seqfree-baselines}
\centering\small\paperTableStyle
\begin{tabular*}{\linewidth}{@{\extracolsep{\fill}}lS[table-format=2.2]S[table-format=2.2]S[table-format=2.2]S[table-format=3.2]@{}}\toprule
 & \multicolumn{2}{c}{Joint accuracy} & & \\
\cmidrule(lr){2-3}
Configuration & {@0.5} & {@0.7} & {mIoU} & {TR} \\
\midrule
Seq-free & 62.00 & 52.40 & 55.14 & 99.53 \\
Router + VEGA & 69.33 & 61.87 & 61.22 & 99.53 \\
TT-VG & 69.80 & 62.33 & 61.52 & 100.00 \\
\bottomrule\end{tabular*}
\par\smallskip\parbox{\linewidth}{\footnotesize Seq-free uses Qwen3-VL-8B; Router denotes Pred-Name Router. TR measures whether the target identity is correctly resolved.}
\end{table}

The three-stage comparison tests the interface downstream of resolution. Routing Seq-free's predicted name to VEGA raises Joint@0.5 to 69.33\%, a 7.33-point increase. TSTT reaches 100.00\% target resolution and 69.80\% Joint@0.5, adding 0.47 points with VEGA held fixed. The larger increment jointly changes identity-to-support routing, selected-support input organization, and learned grounding; the second replaces predicted-name resolution with executable state resolution while holding VEGA fixed. Together they yield a 7.80-point improvement over direct inference, with paired intervals in Table~\ref{tab:corrected-statistics}. Failure and routing analyses locate the dominant observed bottleneck in using selected visual evidence after target resolution.
\par
\begin{table}[pos=!t]
\caption{Paired improvements with 95\% bootstrap intervals. Differences are percentage points; cluster intervals resample episodes grouped by their target base pair.}\label{tab:corrected-statistics}
\centering\small\paperTableStyle
\parbox{\linewidth}{\textbf{A. Seq-free $\to$ Router + VEGA}}
\par\vspace{4pt}
\begin{tabular*}{\linewidth}{@{\extracolsep{\fill}}lrrr@{}}\toprule
Metric & $\Delta$ & Episode CI & Cluster CI \\
\midrule
Joint@0.5 & +7.33 & [4.87, 9.73] & [4.50, 10.07] \\
mIoU & +6.08 & [4.23, 7.90] & [3.88, 8.21] \\
\bottomrule\end{tabular*}
\par\vspace{7pt}
\parbox{\linewidth}{\textbf{B. Router + VEGA $\to$ TT-VG}}
\par\vspace{4pt}
\begin{tabular*}{\linewidth}{@{\extracolsep{\fill}}lrrr@{}}\toprule
Metric & $\Delta$ & Episode CI & Cluster CI \\
\midrule
Joint@0.5 & +0.47 & [0.13, 0.87] & [0.14, 0.81] \\
mIoU & +0.30 & [0.08, 0.58] & [0.08, 0.58] \\
\bottomrule\end{tabular*}
\par\smallskip\parbox{\linewidth}{\footnotesize Router denotes Pred-Name Router; Panel B holds VEGA weights and visual-input organization fixed.}
\end{table}
The selected-support views in Figure~\ref{fig:seqfree-failures} are explanatory; Seq-free uses its original dialogue inputs. The displayed boxes and IoUs come from the retained primary predictions.
Figure~\ref{fig:grounding-diagnostics} (right) retains all predictions while varying the IoU threshold; the primary threshold remains 0.5. The observed TT-VG--Seq-free difference is positive across the plotted thresholds. Appendix~\ref{app:reading-diagnostics} gives exploratory paired uncertainty estimates for the selected IoU thresholds.

\subsubsection{Selected-Evidence Controls}\label{sec:controls}
\begin{figure*}[pos=t]
\centering\includegraphics[width=\textwidth]{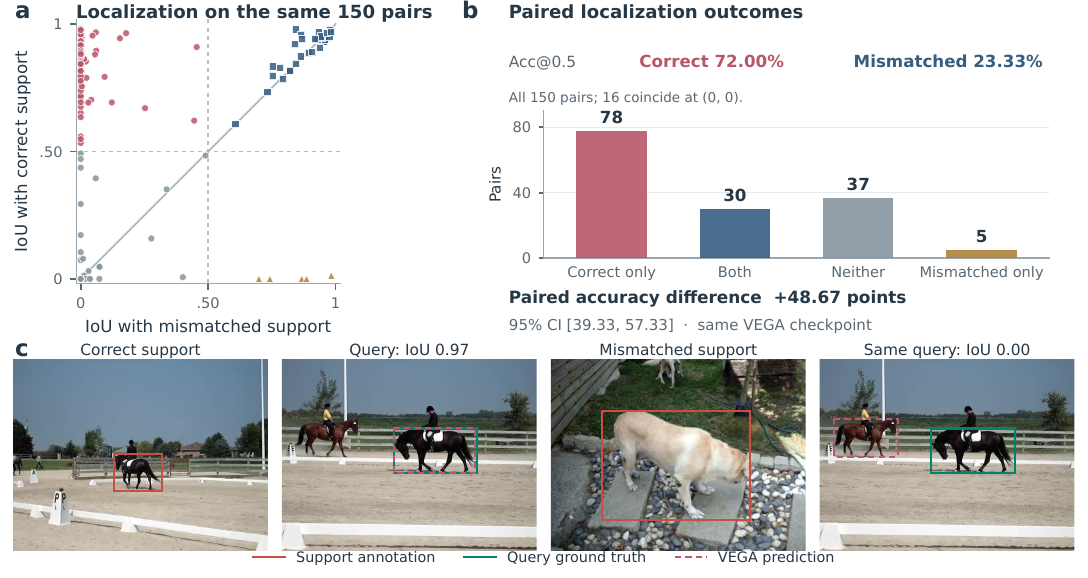}
\caption{Fixed-checkpoint support intervention on 150 pairs. Panels (a,b) show paired IoUs and success counts, with 0.5 thresholds and an equal-IoU diagonal; panel (c) illustrates correct versus mismatched support records, showing the support-context view from each full package (red annotations), query ground truth (solid green), and predictions (dashed rose). Changing the selected support record changes localization on the same query; the displayed example is illustrative.}\label{fig:evidence-controls}
\end{figure*}

On all 1,500 episodes, selecting the latest or a random support reduces Joint@0.5 to about 19\%, with target resolution near 28\% (Table~\ref{tab:additional-baselines}). The VEGA checkpoint and package construction remain fixed across routing conditions; only the selected record changes. Figure~\ref{fig:evidence-controls} examines how changing the supplied support affects localization on identical query images, using paired predictions from a fixed checkpoint.

The fixed-checkpoint pilot selects ten pairs per domain--curated-stratum cell before inference (seed 20261008). Full evidence reaches 72.00\% Acc@0.5, compared with 23.33\% for mismatched support: a 48.67-point paired difference (95\% interval [39.33, 57.33]). The full-minus-crop-only difference is 18.00 points [11.33, 24.67], whereas full minus boxed context is 2.67 [-2.00, 7.33]. The selected support pair strongly affects localization; the added value of the crop over boxed context is estimated less precisely. Appendix~\ref{app:evidence-details} specifies the intervention, and Appendix~\ref{app:qwen-input-control} reports matched input controls.

\subsubsection{Task-Adaptation Controls}\label{sec:backend-adaptation}
Table~\ref{tab:localization-comparison} fixes the TSTT-resolved evidence and target resolution, isolating query-side grounding under each method's stated training condition. VEGA reaches 69.80\% Joint@0.5 and 61.52\% mIoU. It exceeds the strongest frozen backend, GDINO--MASA-R50, by 16.00 Joint@0.5 points. We next test whether VEGA's advantage over frozen retrieval backends persists when those backends receive task adaptation.
We adapt two proposal-based retrieval components using the same 5,000 LaSOT training pairs: the final transformer block and output norm of DINOv2, and the existing embedding projection of MASA. Each receives one epoch of support--query contrastive training. Fresh frozen and adapted conditions share the selected support and retained Grounding DINO proposal pool on all 1,500 episodes; target resolution is 100\% in every condition. The proposal detector stays fixed throughout. Appendix~\ref{app:backend-adaptation} gives the full adaptation settings.

The paired Joint@0.5 gains are +2.40 points for DINOv2 (95\% interval [1.06, 3.84]) and +1.87 for MASA (95\% interval [-0.07, 3.80]).\par

\par

\begin{table}[pos=htbp]
\caption{Paired retrieval-backend adaptation on all 1,500 episodes. Frozen and adapted conditions share selected supports and proposal pools; scores are percentages.}\label{tab:backend-adaptation}
\centering\small\paperTableStyle
\begin{tabular*}{\linewidth}{@{\extracolsep{\fill}}lS[table-format=2.2]S[table-format=2.2]S[table-format=2.2]S[table-format=2.2]@{}}\toprule
& \multicolumn{2}{c}{Joint@0.5} & \multicolumn{2}{c}{mIoU}\\\cmidrule(lr){2-3}\cmidrule(lr){4-5} Backend & {Frozen} & {Adapted} & {Frozen} & {Adapted} \\
\midrule
DINOv2 & 49.07 & 51.47 & 48.08 & 49.68 \\
MASA & 53.80 & 55.67 & 50.07 & 51.70 \\
\bottomrule\end{tabular*}
\end{table}

Intervals use 1,050 target-base-pair clusters and 5,000 bootstrap resamples; aggregation remains episode-weighted. DINOv2 has a positive paired interval; MASA has a positive point estimate with an interval spanning zero. The DINOv2 frozen reference here uses bfloat16 inference (49.07\%), whereas Table~\ref{tab:localization-comparison} uses the retained FP32 reference (49.33\%). Training objectives and trainable scopes are specific to each backend. The retained predictions permit independent rescoring of both conditions.
The adapted backends reach 51.47\% and 55.67\% Joint@0.5, while VEGA reaches 69.80\% under matched target resolution.

\subsubsection{State Restoration and Dependency Analysis}
\begin{figure}[pos=!ht]
\centering\includegraphics[width=\linewidth]{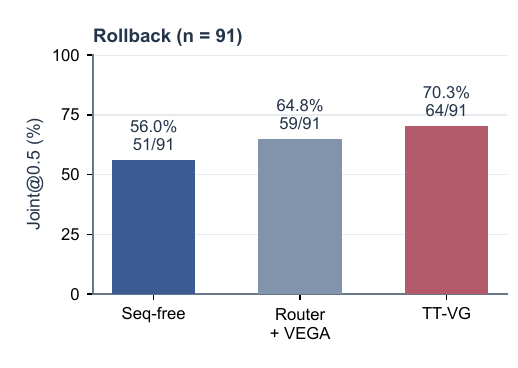}
\caption{Rollback comparison on all 91 episodes. Bars show Joint@0.5 and success counts; Router + VEGA and TT-VG share VEGA weights. Restoration adds five successes (5.49 points); Appendix~\ref{app:dependency-groups} gives all groups.}\label{fig:dependency-groups}
\end{figure}

Figure~\ref{fig:dependency-groups} highlights the state-restoration dependency. With VEGA fixed, TSTT raises Rollback Joint@0.5 from 64.84\% (59/91) to 70.33\% (64/91), a 5.49-point increase. The overall TSTT increment remains 0.47 points; the additional benefit is concentrated in episodes requiring explicit state restoration. Appendix~\ref{app:dependency-groups} reports all four dependencies, with sample sizes and the two stage increments.

\subsection{Controlled Wording Transfer}\label{sec:robustness-summary}
To test sensitivity to wording while preserving state semantics, we evaluate a frozen Qwen3-VL-8B text front end on four paired renderings of 40 fresh episode identifiers. Final-target and full-program agreement are 40/40 for familiar registration/selection wording, 36/40 for catalogue/item-of-interest wording, 33/40 for visual-record/focus wording, and 40/40 for concise commands. Errors concentrate in reference interpretation, indicating sensitivity to how the same state operation is expressed. Full prompts, JSON-validity outcomes, contract-validation, long-state, and paired-state diagnostics are provided in Appendices~\ref{app:parser-iterations}--\ref{app:state-cost}.

\section{Discussion}
Interaction state determines both what the user requests and which stored visual evidence is relevant. SCVIB makes this source-relevance decision executable and measures the chain from target resolution to query-side localization. The controlled protocol supports identifiable state semantics across direct lookup, active targets, non-latest sources, and rollback. This common diagnostic setting supports analysis of instance-specific information selection and use.

The resolution--localization gap shows that successful resolution alone is insufficient. Direct Qwen3-VL-8B reaches 99.53\% target resolution and 62.00\% Joint@0.5. Routing the predicted name to selected support and learned grounding yields the larger improvement in the three-stage comparison. Fixed-checkpoint support substitution reduces Acc@0.5 from 72.00\% to 23.33\%. Together, these observations diagnose a post-resolution evidence-utilization bottleneck: the resolved identity must be connected to relevant appearance and context and used effectively in the query scene. The routing increment jointly changes identity-to-record linking, selected-support input organization, and learned grounding; paired interventions measure localization responses to support changes.

TT-VG provides a state-conditioned visual grounding framework with an explicit and auditable state-to-evidence interface. TSTT traces visible edits and restoration to the selected record. The Visual Evidence Package organizes target appearance, boxed context, and the query for learned support-conditioned grounding by VEGA. TSTT's additional benefit is concentrated on Rollback, consistent with its execution role. Task adaptation improves both retrieval backends, although the paired evidence is clear for DINOv2 and less certain for MASA. For the same resolved targets, both adapted backends retain a localization gap relative to VEGA on the canonical benchmark.

For information fusion, SCVIB highlights source relevance as an interaction-conditioned decision that precedes visual integration. A state edit can change the required support without changing the stored images. An explicit selected-evidence interface makes this decision inspectable and permits controlled interventions on the inputs to grounding. The benchmark and framework provide a basis for studying selective information use in interactive multimodal systems by connecting executable state to the integration of instance appearance, context, and query evidence.

\section{Limitations}
The controlled diagnostic protocol uses closed interaction language, episode-local names, and short histories. Wording-transfer templates broaden lexical coverage within that ontology. Freer independently human-written language, longer multimodal histories, cross-session persistence, and broader personalized baselines and model families provide natural extensions for evaluating generalization. 

The curated strata depend on the construction model and prompt, so stratum-level scores describe this selected sample. Repeated target pairs and shared query images motivate clustered uncertainty estimates; shared source scenes may retain some residual dependence.

The adaptation controls share the same training pairs but retain architecture-specific learning objectives, trainable parameter scopes, and visual preprocessing pipelines. Deployment cost as interaction histories grow beyond the evaluated setting remains an open question.

\section{Conclusion}
Interaction state determines which stored evidence is currently relevant as well as what the user requests, including when a state edit leaves the stored images unchanged. SCVIB operationalizes this state-dependent source relevance as a controlled diagnostic benchmark and reveals a 37.53-point resolution--localization gap: direct inference resolves targets with 99.53\% accuracy but reaches 62.00\% Joint@0.5. Failure decomposition, routing comparisons, and support interventions diagnose a post-resolution evidence-utilization bottleneck. The TT-VG framework makes the state-to-evidence-to-grounding path explicit and auditable. It combines TSTT's execution of visible edits and state restoration with the selection of relevant support records and VEGA's learned support-conditioned grounding, reaching 69.80\% Joint@0.5. The findings connect interactive state semantics to selective information fusion: correct resolution alone is insufficient; relevant visual evidence must be selected and connected to the requested target, then effectively utilized to localize the same instance in the query image.

\section*{CRediT authorship contribution statement}
\noindent\textbf{Xiongtai Yang:} Conceptualization, Software, Data curation, Writing -- original draft, Writing -- review \& editing, Validation.\par
\noindent\textbf{Ziyan He:} Conceptualization, Writing -- original draft.\par
\noindent\textbf{Tao Wang:} Conceptualization, Validation, Supervision.

\section*{Declaration of competing interests}
The authors declare that they have no known competing financial interests or personal relationships that could have appeared to influence the work reported in this paper.

\section*{Funding}
This research did not receive any specific grant from funding agencies in the public, commercial, or not-for-profit sectors.

\section*{Data availability}
Evaluation code is provided as Supplementary Material with the journal submission. Upon acceptance, SCVIB annotations, interaction manifests, source metadata, hashes, and reconstruction utilities will be made available. Image access will follow the upstream dataset licenses through permitted redistribution or reproducible acquisition and reconstruction from the original sources. Appendix~\ref{app:source-access} specifies the canonical source inventory and reconstruction procedures.
\FloatBarrier
\appendix
\numberwithin{table}{section}
\numberwithin{figure}{section}
\numberwithin{equation}{section}
\section{Executable Interface and Reproducibility}\label{app:reproducibility}
\setcounter{dbltopnumber}{2}
The executable interface uses episode and rendering identifiers, chronological visible text, and support-image references. Evaluation annotations---ground-truth names and boxes, target-turn identifiers, and selected-support indices---are separate from compiler inputs.

\subsection{Controlled event serialization}
Each state version stores the active name, introduced-name set, predecessor identifier, and switch-undo stack depth (Table~\ref{tab:appendix-events}). Set/mark operations leave the undo stack unchanged, including between a switch and its undo.

\begin{table}[pos=htbp]
\caption{Canonical executor events and their effects. The registry is append-only; rollback retains all introduced evidence.}\label{tab:appendix-events}
\centering\footnotesize\paperTableStyle
\begin{tabular*}{\linewidth}{@{\extracolsep{\fill}}ll@{}}\toprule
Event & Effect \\
\midrule
\texttt{introduce\_only} & Register; active unchanged. \\
\texttt{set\_active} & Register; set active. \\
\texttt{keep\_active} & Register distractor; active unchanged. \\
\texttt{switch\_active} & Register; save active; switch. \\
\texttt{mark\_active} & Set active to a registered name. \\
\texttt{undo\_previous\_switch} & Pop switch; restore active. \\
\bottomrule\end{tabular*}
\end{table}

\begin{figure*}[pos=t]
\centering
\includegraphics[width=\textwidth]{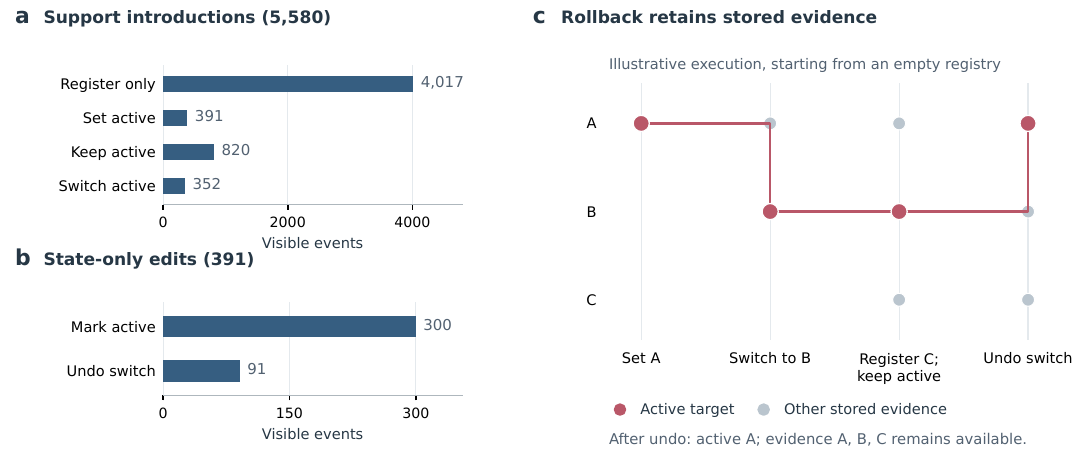}
\caption{Visible-event distribution and an illustrative rollback execution. Panels (a,b) count compiled events in 1,500 histories on separate axes; panel (c) tracks the active target and available evidence through four edits. Undo restores the active target while preserving subsequently registered evidence.}\label{fig:event-inventory}
\end{figure*}

The visible histories contain 5,580 support introductions and 391 state-only edits. Figure~\ref{fig:event-inventory} summarizes the emitted events, distinguishing support-associated updates from text-only state changes.

Final references select a direct name, the current active target, the earliest introduction, or the predecessor of a named anchor in support-introduction order. Predecessor lookup uses registration order, independently of activation and switch order. Anchors without predecessors are rejected.

\subsection{Worked rollback trace}
For the executed history in Figure~\ref{fig:event-inventory}c, current-active lookup returns A, direct-name lookup for C returns C, and predecessor-of-C lookup returns B. Each lookup leaves the executed state unchanged.

\subsection{Multimodal prompt templates}
Table~\ref{tab:prompt-interfaces} summarizes the three prompt interfaces. In Qwen user messages, images precede visible text; text-only events retain their chronological positions. Direct-inference acknowledgements are generated assistant messages. The State-Table summary comes from deterministic visible-state execution and adds introduction order, active names, and introduced-name sets; it includes neither the compiled final reference nor its resolved evidence.

\begin{table}[pos=htbp]
\caption{Prompt interfaces and returned fields. Name references in the direct conditions retain the canonical name, active-target, or support-order instruction.}\label{tab:prompt-interfaces}
\centering\footnotesize\paperTableStyle
\begin{tabularx}{\linewidth}{@{}p{.23\linewidth}X@{}}\toprule
Interface & Model-visible input and required output \\
\midrule
Seq-free & Ordered support/update dialogue and final query; return one name-and-box JSON object. \\
State-Table & Same dialogue plus introduction order and a linearized state summary; return name and box. \\
VEGA & Target crop, full support with a red target box, then query; return \texttt{\{"bbox\_\allowbreak{}2d":[x1,y1,x2,y2]\}}. No pseudo-name is supplied. \\
\bottomrule\end{tabularx}
\end{table}

VEGA's fixed instruction identifies the first image as target appearance and requests instance-discriminative color, texture, pattern, accessories, and other details. This instruction is held fixed in the full-package control. Crop padding is 3\% for training and the original single-turn/fixed-checkpoint pilots, and 0\% for the main multi-turn, matched-noise, and matched-adapter evaluations.

\subsection{Parsing and coordinate conventions}
The primary Seq-free and State-Table parser prefers the final \texttt{<answer>} block, removes Markdown fences, and tries balanced JSON objects or arrays. Fallbacks use explicitly named name/box fields or the last bracketed four-number box. Both \texttt{bbox} and \texttt{bbox\_\allowbreak{}2d} are accepted. Names match exactly after case and surrounding-quote normalization, without fuzzy aliases. Table~\ref{tab:coordinate-conventions} records the method-specific coordinate rules.

\begin{table}[pos=htbp]
\caption{Coordinate handling and invalid-output rules. Proposal and mask outputs are normalized using query-image dimensions.}\label{tab:coordinate-conventions}
\centering\footnotesize\paperTableStyle
\begin{tabularx}{\linewidth}{@{}p{.27\linewidth}X@{}}\toprule
Interface & Convention \\
\midrule
Seq-free / State-Table & Swap reversed endpoints, then clip to $[0,1000]$. Missing or non-positive-area boxes receive zero IoU. \\
VEGA parser & Clip to $[0,1000]$ without swapping endpoints; invalid boxes receive zero IoU. \\
Single-turn input & Clip support/query annotations to pixel-index bounds $w-1,h-1$ before drawing, cropping, and serializing supervision. \\
Multi-turn scoring & Use the accompanying continuous query annotations. \\
\bottomrule\end{tabularx}
\end{table}

\subsection{Localization interfaces}
Table~\ref{tab:supp-localization-interfaces} specifies the four localization interfaces. Proposal and mask outputs follow each backend's native pixel-index convention before normalization.

\begin{table*}[pos=t]
\caption{Query-side localization interfaces. Every method receives the same TSTT-resolved support record and query image, fixing target resolution at 100.00\%.}\label{tab:supp-localization-interfaces}
\centering\footnotesize\paperTableStyle
\begin{tabular*}{\linewidth}{@{\extracolsep{\fill}}lll@{}}\toprule
Localization method & Support representation & Query-side procedure \\
\midrule
GDINO--DINOv2 & Target crop & Grounding DINO proposals; DINOv2 similarity \\
GDINO--MASA-R50 & Selected support & Grounding DINO proposals; MASA-R50 matching \\
Box-initialized PerSAM & Support box & Personalized segmentation; mask-to-box \\
VEGA & Visual Evidence Package & VEGA support-conditioned grounding \\
\bottomrule\end{tabular*}
\end{table*}

\paragraph{GDINO--DINOv2.}
GDINO--DINOv2 uses the fixed proposal vocabulary below and the settings in Table~\ref{tab:supp-gdino-config}~\citep{Liu2025GroundingDINO,Oquab2023DINOv2}.

\paragraph{GDINO--MASA-R50.}
The released MASA-R50 model ranks the shared frozen proposals by cosine similarity of L2-normalized support and proposal embeddings~\citep{Li2024MASA}. The highest-ranked proposal supplies the query box in its original coordinates.

\paragraph{Box-initialized PerSAM.}
The released Personalize-SAM implementation uses SAM ViT-H and the resolved support box as its support-side prompt~\citep{Zhang2024PerSAM}. The resulting target representation guides query-mask prediction; the tight box of the final non-empty mask is returned. Query ground truth is used only for scoring.

\begin{table}[pos=htbp]
\caption{GDINO--DINOv2 backend configuration. The full proposal vocabulary is specified alongside the table.}\label{tab:supp-gdino-config}
\centering\footnotesize\paperTableStyle
\begin{tabular*}{\linewidth}{@{\extracolsep{\fill}}ll@{}}\toprule
Parameter & Setting \\
\midrule
Proposal model & GroundingDINO Swin-T OGC \\
Box / text threshold & 0.05 / 0.05 \\
Maximum proposals & 100 \\
Class-agnostic NMS IoU & 0.70 \\
Support / proposal crop margin & 0.10 \\
Matcher & DINOv2 ViT-B/14 + 4 registers \\
Feature / similarity & CLS / cosine similarity \\
SCVIB-specific training & None \\
\bottomrule\end{tabular*}
\end{table}
\paragraph{Fixed proposal vocabulary.}
\texttt{object. person. animal. vehicle. car. truck. bus. bicycle. motorcycle. furniture. animated character.}

\subsection{VEGA Training Configuration}\label{app:vega-training}
VEGA adapts Qwen3-VL-4B-Instruct using 5,000 same-instance pairs derived from LaSOT trajectories across 64 sequences. The 164 LaSOT evaluation sequences are disjoint (Appendix~\ref{app:source-access}). Table~\ref{tab:vega-recipe} collects the training and inference settings.
\begin{table}[pos=htbp]
\caption{Canonical VEGA adaptation and inference recipe. All routing conditions share the same checkpoint and Visual Evidence Package.}\label{tab:vega-recipe}
\centering\footnotesize\paperTableStyle
\begin{tabularx}{\linewidth}{@{}p{.29\linewidth}X@{}}\toprule
Setting & Value \\
\midrule
LoRA & Rank 16; $\alpha=32$; dropout 0.05. \\
Adapted projections & Attention \texttt{q/k/v/o}; MLP \texttt{gate/up/down}. \\
Supervision & Answer-token cross-entropy on the serialized normalized query box; prompt/padding tokens masked. \\
Optimizer & AdamW; learning rate $10^{-4}$; zero weight decay; gradient clipping 1. \\
Training schedule & One epoch; batch size 1; gradient accumulation 8. \\
Seed / precision & $s=3$ in Python, PyTorch and CUDA; bfloat16. \\
Visual input & Target crop, boxed support context, query; at most 401,408 pixels per image. \\
Support crop & Clipped support box with 3\% padding in training; tight crop in multi-turn inference. \\
Grounding decoding & Greedy, capped at 64 tokens. \\
\bottomrule\end{tabularx}
\end{table}

\section{Evaluation and Stratum Calibration}\label{app:historical-audit}
All final results use the canonical clean-source input version, \texttt{scvib-clean-source-v1} (Appendix~\ref{app:source-access}). Source cleaning replaces 102 overlay-bearing frames from 51 pairs with verified source frames. Configurations affected by source cleaning or final-reference normalization are reevaluated on their affected inputs with fixed checkpoints and support-rendering rules. Unaffected predictions are reused. Proposal-based retrieval comparisons share Grounding DINO proposals for the cleaned queries.

Direct inference retains visible dialogue prefixes, generated acknowledgements, and the canonical final-reference instruction (Table~\ref{tab:prompt-interfaces}). Pred-Name Router uses the fixed VEGA checkpoint and package; outputs are reused only when the selected visual evidence supplied to the grounder is identical.

The main evaluator uses continuous multi-turn query annotations. Single-turn preparation clips annotations to $[0,w-1]\times[0,h-1]$; 71 of 1,050 query boxes differ at a boundary. These single-turn results use the clipped annotation convention. Parsing and invalid-output handling follow Table~\ref{tab:coordinate-conventions}.

File hashing identifies 987 distinct query-image files across 1,050 pairs, with some scenes shared across pairs. A shared-query-image sensitivity check clusters episodes by query-file hash and gives Joint@0.5 delta intervals [4.45, 10.15] for Seq-free to predicted-name routing and [0.14, 0.86] for the Router-to-TT-VG increment.

Stratum calibration uses construction-time localization predictions aligned to the support/query media and annotations used for curation. Of 1,043 compatible predictions, 970 agree with the curated stratum and 73 disagree with the intended IoU thresholds. Seven further pairs lack compatible construction-time predictions. Benchmark metadata distinguishes the curated label, threshold assignment, and missing or ambiguous status. The curated split remains 70 pairs per domain--stratum cell; recalibrated assignments may yield unequal group sizes. Alternative construction prompts also change assignments on the real-image subset.

\section{Source Inventory and Reconstruction}\label{app:source-access}
\subsection{Source-access inventory and descriptive sensitivity}
The inventory identifies sources, sequences/videos, support/query frames, and annotation records for all 1,050 pairs. The external sources are LaSOT, MOT17, BDD100K, Argoverse, YFCC100M, and Charades; 210 Anime-Style pairs were generated using GPT Image 2 (OpenAI; \texttt{gpt-image-2}), selected, and manually annotated as support--query pairs. LaSOT, MOTChallenge, and the nested source collections are documented by their upstream projects~\citep{Fan2019LaSOT,Dendorfer2021MOTChallenge,Thomee2016YFCC100M,Chang2019Argoverse,Yu2020BDD100K,Sigurdsson2016Charades}. PDM and TAO are acquisition or annotation collections rather than additional mutually exclusive image sources. The nested-source inventory identifies 392 TAO-derived and 112 MOT17-derived pairs. All 271 LaSOT evaluation pairs span 164 sequences, disjoint from the 64 training sequences.

\subsection{Canonical image reconstruction}
The canonical version, \texttt{scvib-clean-source-v1}, covers 1,050 pairs and 2,100 image records. Reconstruction starts from hash-verified upstream frames and author-generated Anime-Style assets.

\paragraph{Direct source copy (1,702 records).}
Verified source bytes are copied directly, including the 102 clean frames replacing review overlays.

\paragraph{Deterministic encoding (398 records).}
The remaining 199 directly collected LaSOT/MOT17 pairs undergo two RGB-decode/JPEG-encode passes: quality 95, 4:2:0 subsampling, no optimization or progressive encoding. The encoder uses Pillow 12.1.1 and libjpeg-turbo 3.1.3. All 398 outputs match their frozen SHA-256 hashes; verification covers all 2,100 records and 987 distinct query hashes.

Media access follows the Data availability statement; pretrained checkpoints are obtained under their respective upstream terms. Table~\ref{tab:source-access} reports descriptive clean-input scores by source, with the original benchmark curation retained.
\par
\begin{table}[pos=htbp]
\caption{Descriptive Joint@0.5 (\%) by resolved underlying source. Underlying image sources and acquisition collections are distinguished in the inventory.}\label{tab:source-access}
\centering\footnotesize\paperTableStyle\footnotesize\begin{tabular*}{\linewidth}{@{\extracolsep{\fill}}lrrrr@{}}\toprule
Source & Pairs & Episodes & Seq-free & TT-VG\\\midrule
Anime-Style & 210 & 300 & 49.67 & 55.67 \\
LaSOT & 271 & 381 & 71.13 & 79.53 \\
YFCC100M & 147 & 212 & 65.57 & 74.06 \\
Argoverse & 86 & 127 & 62.20 & 74.80 \\
BDD100K & 99 & 141 & 49.65 & 63.12 \\
Charades & 27 & 39 & 84.62 & 94.87 \\
MOT17 & 210 & 300 & 63.00 & 66.33 \\
\bottomrule\end{tabular*}\end{table}

\section{Additional Baselines and Evidence-Control Details}\label{app:extended-controls}
Table~\ref{tab:additional-baselines} adds the State-Table interface from Table~\ref{tab:prompt-interfaces}, retaining the visual dialogue. The latest/random selectors hold VEGA fixed; frozen 4B inference matches VEGA's base-model size.
\par
\begin{table}[pos=hbp]
\caption{Additional configurations evaluated on all 1,500 episodes. All reported scores are percentages.}\label{tab:additional-baselines}
\centering\footnotesize\paperTableStyle\footnotesize
\begin{tabular*}{\linewidth}{@{\extracolsep{\fill}}lrrrr@{}}\toprule
Configuration & Joint@0.5 & Joint@0.7 & mIoU & TR \\
\midrule
Visible State-Table & 58.13 & 48.20 & 51.98 & 99.40 \\
Latest-Support + VEGA & 19.07 & 16.33 & 30.52 & 28.00 \\
Random-Support + VEGA & 18.87 & 17.13 & 29.76 & 28.27 \\
Seq-free (4B) & 48.93 & 42.87 & 46.31 & 97.60 \\
\bottomrule\end{tabular*}\end{table}

\subsection{Fixed-checkpoint evidence specification}\label{app:evidence-details}
On the preselected 150-pair set in Section~\ref{sec:controls}, donors are rotated within each domain--curated-stratum cell, excluding sample self-assignment without enforcing donor--recipient identity correspondence. Image-removal conditions adjust instructions to the remaining image roles. The fixed-checkpoint pilot uses 3\% crop padding; the canonical multi-turn package uses a tight crop.
Table~\ref{tab:fixed-evidence} gives all four conditions; Section~\ref{sec:controls} reports the paired accuracy contrasts and intervals.
\begin{table}[pos=htbp]
\caption{Fixed-checkpoint evidence controls on the same 150 pairs. Accuracy and mIoU are percentages; all conditions retain the full denominator.}\label{tab:fixed-evidence}
\centering\footnotesize\paperTableStyle
\begin{tabular*}{\linewidth}{@{\extracolsep{\fill}}lrrr@{}}\toprule
Evidence & Acc@0.5 & Acc@0.7 & mIoU \\
\midrule
Full & 72.00 & 64.00 & 63.83 \\
Boxed context & 69.33 & 62.00 & 61.09 \\
Crop & 54.00 & 48.00 & 48.13 \\
Mismatched support & 23.33 & 22.00 & 23.02 \\
\bottomrule\end{tabular*}
\end{table}

\subsection{Qwen input-organization control}\label{app:qwen-input-control}
\begin{table}[pos=htbp]
\caption{Matched-training-recipe pilot comparing two adaptations of Qwen3-VL-4B on 150 fixed pairs with correctly selected supports. These are grounding scores; the adaptations share training examples and optimizer settings, with distinct visual inputs and associated instructions.}\label{tab:matched-adapter}
\centering\footnotesize\paperTableStyle\footnotesize\begin{tabular*}{\linewidth}{@{\extracolsep{\fill}}lrr@{}}\toprule
Measure & Two-image Qwen & VEGA \\
\midrule
Acc@0.5 (\%) & 72.00 & 72.00 \\
Acc@0.7 (\%) & 62.00 & 64.00 \\
mIoU (\%) & 61.99 & 64.03 \\
\bottomrule\end{tabular*}\end{table}
\begin{table*}[pos=t]
\caption{Backend-specific adaptation settings. Visual preprocessing and trainable components follow the respective retrieval architectures.}\label{tab:retrieval-recipes}
\centering\footnotesize\paperTableStyle
\begin{tabularx}{\linewidth}{@{}p{.19\linewidth}XX@{}}\toprule
Setting & DINOv2 & MASA \\
\midrule
Backbone & ViT-B/14 with four register tokens & R50 \\
Updated component & Final transformer block and output norm & Existing 1024-to-256 embedding projection \\
Trainable parameters & 7,090,944 & 262,400 \\
Learning rate / decay & $10^{-5}$ / 0.04 & $10^{-4}$ / 0 \\
Visual preprocessing & Support/candidate crops with 10\% margin; $224\times224$ bicubic resize; ImageNet normalization & Full images with maximum side 1024 pixels; original ROI feature extraction \\
Computation & bfloat16 autocast in training and paired inference & Frozen upstream features cached for training \\
\bottomrule\end{tabularx}
\end{table*}
We compare two Qwen3-VL-4B-Instruct LoRA adapters using the shared optimization and LoRA settings in Table~\ref{tab:vega-recipe} ($s=3$), with 33,030,144 trainable parameters each. One receives boxed support context and query; VEGA additionally receives the target crop. The adapters share the same 5,000 training pairs, target fields, example order, and support/query paths. Evaluation uses correctly selected supports on the same 150 pairs, byte-identical boxed context and query, tight VEGA crops, and greedy decoding capped at 64 tokens.
\par

The paired VEGA-minus-two-image-Qwen Acc@0.5 difference is 0.00 points (95\% stratified base-pair bootstrap interval [-4.00, 4.00]); the mIoU difference is +2.04 points [-1.13, 5.23]. Appendix~\ref{app:state-cost} gives the deployment environment and measurements.

\subsection{Retrieval-Backend Adaptation}\label{app:backend-adaptation}
Both retrieval backends use VEGA's training set (Appendix~\ref{app:vega-training}). Their shared recipe uses seed $s=3$, one shuffled epoch, batch size 32, 157 AdamW steps, gradient-norm clipping at 1, and symmetric support--query InfoNCE with temperature 0.1. Same-sequence off-diagonal examples are masked as negatives. The final epoch checkpoint is fixed before benchmark evaluation.

The 1,500 episodes use visible-protocol-selected supports and a shared clean-input Grounding DINO pool of 1,050 base-pair queries (987 distinct file hashes), with identical candidate ordering across frozen/adapted conditions. Inference records contain support boxes and query dimensions; query annotations are reserved for scoring. The paired DINOv2 reference uses bfloat16, whereas Table~\ref{tab:localization-comparison} uses FP32; Table~\ref{tab:backend-adaptation} gives the paired reference scores. Invalid outputs remain in the denominator. Percentile intervals use 5,000 paired target-base-pair-cluster resamples (seed 20261010), retaining all episodes of each sampled pair and aggregating over episodes.

\subsection{Clean-input single-turn configurations}
Table~\ref{tab:single-turn-configurations} reports the single-turn configurations. Fixed-checkpoint evidence controls and matched input comparisons appear in Appendices~\ref{app:evidence-details} and~\ref{app:qwen-input-control}, respectively.
\begin{table}[pos=htbp]
\caption{Clean-input single-turn results across distinct training and input configurations. Context denotes boxed support; the crop-adapter row uses shuffled crops. Accuracy is in percent; VEGA uses 1,050 pairs.}\label{tab:single-turn-configurations}
\centering\footnotesize\paperTableStyle
\begin{tabularx}{\linewidth}{@{}p{.26\linewidth}Xr@{}}\toprule
Input & Configuration & Acc@0.5 \\
\midrule
Context + query & Frozen base model & 29.05 \\
Context + query & Trained context adapter & 69.33 \\
Crop + query & Frozen base model & 53.24 \\
Crop + query & Trained crop adapter; shuffled crops & 16.29 \\
Full package & VEGA & 70.29 \\
\bottomrule\end{tabularx}
\end{table}

\section{Parsing and Contract Diagnostics}\label{app:parser-iterations}
A frozen Qwen3-VL-8B front end maps authored visible instructions to event/reference JSON for deterministic execution, withholding expected targets and image annotations. Study samples appear in Table~\ref{tab:parser-study-design} and parsing outcomes in Figure~\ref{fig:frontend-summary}.
\begin{table}[pos=htbp]
\caption{Text-front-end sampling designs. Dependencies are Explicit-Name / Active-Target / Counter-Recency / Rollback; IDs are disjoint across studies.}\label{tab:parser-study-design}
\centering\footnotesize\paperTableStyle
\begin{tabular*}{\linewidth}{@{\extracolsep{\fill}}lrrl@{}}\toprule
Study & IDs & Texts/ID & Dependencies \\
\midrule
Initial wording & 80 & 1 & 32 / 14 / 25 / 9 \\
Prompt revision & 80 & 1 & 32 / 8 / 35 / 5 \\
Wording transfer & 40 & 4 & 16 / 6 / 16 / 2 \\
Clarification & 20 & 4 & 8 / 3 / 9 / 0 \\
\bottomrule\end{tabular*}
\par\smallskip\parbox{\linewidth}{\footnotesize Texts denote alternate renderings. Sampling seeds: revision 20261010, transfer 20261011, clarification 20261013. Transfer covers 39 base pairs; clarification covers 20. Prompts, templates and IDs are fixed before each comparison.}
\end{table}
\begingroup
\makeatletter
\def\@dbflt#1{\@ifnextchar[{\@xdblfloat{#1}}{\@xdblfloat{#1}[t]}}
\makeatother
\begin{figure*}[pos=t]
\centering\includegraphics[width=\textwidth]{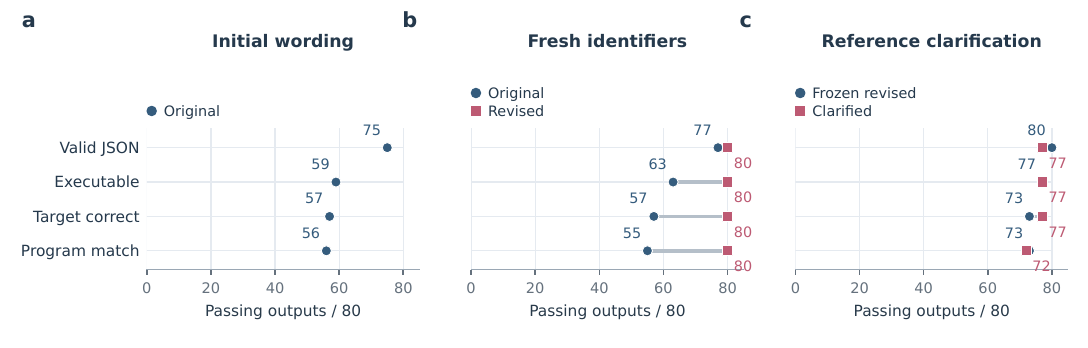}
\caption{Text-front-end diagnostics on three separate sets. Each panel reports four passing-output counts out of 80; connected markers compare prompts on identical inputs. The initial and fresh-identifier sets contain 80 identifiers each, while clarification uses 20 further episodes with four renderings each. Revision improves the fresh-identifier scores; clarification has mixed effects.}\label{fig:frontend-summary}
\end{figure*}
\endgroup

\subsection{Initial diagnostic and long-state execution}
Independent simulation and TSTT agree on all 90 valid closed-grammar traces (30 each at 10, 30, or 100 support introductions), including repeated switches and undo operations.

Initial episodes are randomly sampled across construction templates; introductions, selections, undo operations, and final references are rewritten outside the compiler grammar (Figure~\ref{fig:frontend-summary}(a)). Program agreement merges operationally equivalent bind/keep registrations; correct final names can coexist with event errors. Unit checks cover future-name activation, empty undo history, and unsupported updates.

\subsection{Fresh-identifier paired-prompt comparison}
The revised prompt makes registration explicit for activate/keep/switch, requires one event per instruction, and adds two schematic demonstrations with non-benchmark names. The paired test holds wording families, fresh visible inputs, greedy decoding, and token limit fixed (Figure~\ref{fig:frontend-summary}(b)). Revision corrects 23 targets without adding target errors: +28.75 points (95\% interval [21.25, 36.25]; 10,000 paired bootstrap draws stratified by template).

\subsection{Frozen-Prompt Wording Transfer}\label{app:wording-transfer}
The frozen revised prompt uses the four wording families and agreement measures in Section~\ref{sec:robustness-summary}; renderings preserve names, operations, and final references. JSON is valid except for one Counter-Recency focus rendering (39/40 valid for focus). Ten further errors confuse explicit names with active or predecessor references.

\subsection{Reference-clarification diagnostic}\label{app:reference-clarification}
The follow-up prompt distinguishes explicit names, registration predecessors, current selection, and earliest registration. Against the frozen revised prompt, it corrects seven renderings and introduces three JSON failures on the same four wording families and fresh identifiers (all 80 renderings retained; Figure~\ref{fig:frontend-summary}(c)). The paired target gain is 5.00 points (95\% source-episode-cluster interval [-6.25, 15.00]; 10,000 resamples, seed 20261014).

Five focus renderings return correct targets through direct-name references instead of requested current-active references. Target accuracy rises from 91.25\% to 96.25\%, while full-program agreement falls from 73/80 to 72/80. With this mixed outcome and a target interval spanning zero, clarification remains diagnostic; the main benchmark uses canonical TSTT.

\subsection{Malformed interaction checks}
We construct 280 malformed histories and 280 matched valid controls: 14 contract categories with 20 distinct identifiers each. Categories cover missing active state, future-name activation, empty undo, contradictory switch origin, duplicate registration, absent or first-position predecessor anchors, mixed or repeated final references, unsupported events, duplicate indices, and missing or duplicate final turns. Valid controls have design-specified targets; no image model is invoked. A first-occurrence-only validator accepts 40 repeated-reference cases; enumerating all occurrences and requiring exactly one rejects them. The final validator rejects all 280 malformed histories, resolves all 280 controls, and preserves all 1,500 benchmark resolutions.

\section{State Interventions and Deployment Costs}\label{app:state-cost}
\subsection{Paired state interventions}
Figure~\ref{fig:paired-state-examples} compares histories with identical support images, support order, query image, and final active-target request; only one state-only instruction changes.
\begingroup
\makeatletter
\def\@dbflt#1{\@ifnextchar[{\@xdblfloat{#1}}{\@xdblfloat{#1}[t]}}
\makeatother
\begin{figure*}[pos=t]
\centering\includegraphics[width=\textwidth]{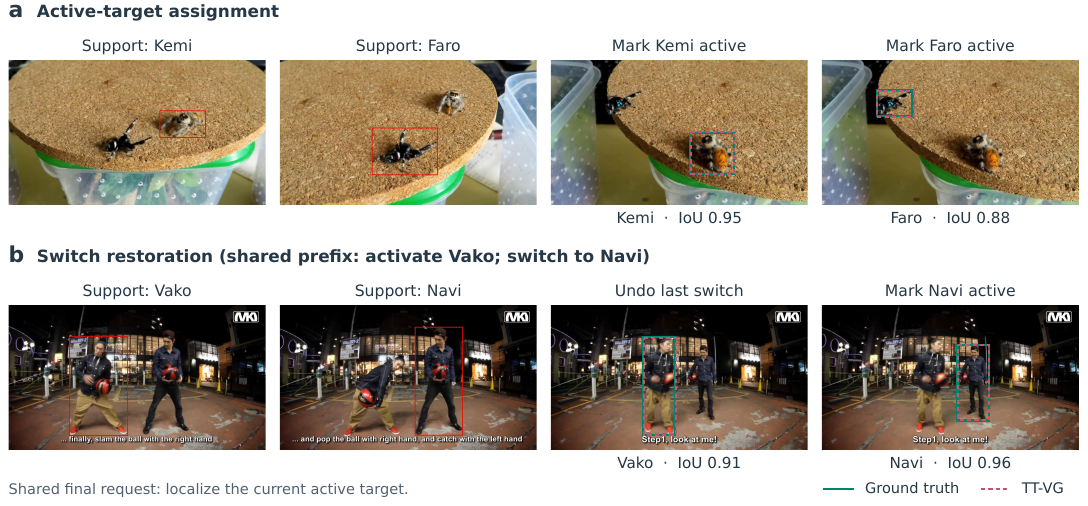}
\caption{State edits on unchanged visual inputs. Each row fixes two supports and a query while varying one instruction; red boxes mark support annotations, solid green is query ground truth, and dashed rose is TT-VG. These selected examples illustrate active assignment and switch restoration; Table~\ref{tab:paired-state} reports all 40 histories.}\label{fig:paired-state-examples}
\end{figure*}
\endgroup

\paragraph{Scene selection.}
From 41 eligible byte-identical query-image groups fixed before analysis, we select one target pair with query-box IoU below 0.1 per group. Shuffling with seed 20261017 fixes the first 20 without using model outcomes: ten active-assignment pairs and ten switch-undo pairs.

\paragraph{Shared inputs.}
The two support acknowledgements are generated once per scene and shared across branches; downstream state acknowledgements are generated separately. Support images and tight crops use the same renderer. Query boxes remain in evaluation annotations and are not read by inference.

\par

\begin{table}[pos=htbp]
\caption{Paired state-response diagnostic on 40 histories from 20 scene pairs. Pair-level metrics require correct predictions for both histories belonging to the same scene.}\label{tab:paired-state}
\centering\footnotesize\paperTableStyle
\begin{tabular*}{\linewidth}{@{\extracolsep{\fill}}lrr@{}}\toprule
Measure & Frozen 4B & TT-VG \\
\midrule
\multicolumn{3}{@{}l}{\textit{History-level evaluation}} \\
Target correct & 40/40 & 40/40 \\
Joint@0.5 (\%) & 42.50 & 72.50 \\
mIoU (\%) & 35.96 & 63.15 \\
\addlinespace[4pt]
\multicolumn{3}{@{}l}{\textit{Pair-level evaluation (both branches)}} \\
Target correct & 20/20 & 20/20 \\
Joint@0.5 correct & 2/20 & 11/20 \\
\bottomrule\end{tabular*}
\end{table}

The frozen Qwen3-VL-4B model changes its predicted name correctly in every pair, demonstrating responsiveness to the controlled state edits. The grounding comparison uses the stated training conditions and generation budgets: task-trained VEGA with 64 grounding tokens versus frozen direct inference with 128 final tokens.

\subsection{Deployment measurements}
Table~\ref{tab:independent-cost} compares two RTX 4090 protocols. The 30-pair run fixes two pairs per domain--curated-stratum cell from the adapter pilot and tests conditions sequentially in separate single-adapter processes. After two excluded warm-ups per condition, caches are cleared and CUDA peak counters reset before each scored call. The 150-pair run keeps both adapters resident and alternates conditions after eight excluded warm-ups.
\par
\begin{table}[pos=htbp]
\caption{Deployment measurements under two protocols. Ratios are median paired VEGA/two-image Qwen times. Single-adapter peaks measure process allocation (base model, adapter and activations), rather than total-device occupancy.}
\label{tab:independent-cost}
\centering\footnotesize\paperTableStyle
\begin{tabular*}{\linewidth}{@{\extracolsep{\fill}}lrr@{}}\toprule
Measure & Two-image Qwen & VEGA \\
\midrule
\multicolumn{3}{@{}l}{\textit{30 pairs: one adapter per process}} \\
Median image tokens & 728 & 803 \\
Median time (s) & 1.814 & 1.890 \\
Median peak allocated (GiB) & 8.730 & 8.769 \\
Maximum peak allocated (GiB) & 8.741 & 8.858 \\
Maximum peak reserved (GiB) & 8.797 & 8.957 \\
Paired time ratio & \multicolumn{2}{r@{}}{1.026} \\
\addlinespace[4pt]
\multicolumn{3}{@{}l}{\textit{150 pairs: both adapters resident}} \\
Median image tokens & 728 & 805 \\
Median time (s) & 1.787 & 1.822 \\
Paired time ratio & \multicolumn{2}{r@{}}{1.013} \\
\bottomrule\end{tabular*}
\end{table}
Both timings cover preprocessing and grounding generation, excluding model loading, image hashing, state execution and dialogue acknowledgements. Additional exclusions are cache clearing (30 pairs) or adapter activation (150 pairs). The latter run uses PyTorch 2.8.0, Transformers 4.57.1 and PEFT 0.17.1, greedy decoding capped at 64 tokens, and 401,408 pixels per image. Support resolution is 150/150 and all 300 boxes are valid. For each image pair, we divide VEGA's elapsed time by the two-image Qwen time, then report the median of these ratios. This paired statistic differs from dividing the two marginal median times in Table~\ref{tab:independent-cost}.

CPU compilation and resolution of 40 histories (100 repetitions each) takes a median 0.060 ms, excluding I/O and cold-process startup. Deployment costs are comparable; VEGA uses modestly more visual tokens and memory.

\setcounter{dbltopnumber}{2}
\renewcommand{\dbltopfraction}{0.85}
\section{Visual and Subgroup Diagnostics}\label{app:extra-diagnostics}
\raggedbottom
\subsection{Support-box perturbations}
The matched support-box study reuses the 150 base-pair IDs fixed before evidence-control inference, selecting the first episode per pair by identifier. It uses tight crops, separately from the 10\% padding diagnostic. Each box edge receives an independent uniform offset bounded by $r$ times the original box width or height, for $r\in\{0,0.05,0.10,0.20\}$. A fixed random direction per pair is scaled across nonzero levels (seed 20261009). After clipping to support-image bounds, the tight crop and boxed context are regenerated with the same renderer and encoding at every level, including the fresh clean baseline. Selected support, query, checkpoint, prompt, greedy decoding and image-token budget remain fixed. The 10,000 paired bootstrap draws preserve all 15 domain--curated-stratum cells; invalid outputs remain in the denominator.
Figure~\ref{fig:matched-noise} gives exact accuracy values and paired drops with 95\% intervals. At 0\%, 5\%, 10\%, and 20\% jitter, mIoU is 63.81\%, 60.56\%, 60.05\%, and 60.93\%, respectively. The accuracy response is non-monotonic: the 5\% and 10\% drop intervals exclude zero, while the 20\% interval includes it. These exploratory intervals are unadjusted for the three contrasts.

\begingroup
\makeatletter
\def\@dbflt#1{\@ifnextchar[{\@xdblfloat{#1}}{\@xdblfloat{#1}[t]}}
\makeatother
\begin{figure*}[pos=t]
\centering\includegraphics[width=\textwidth]{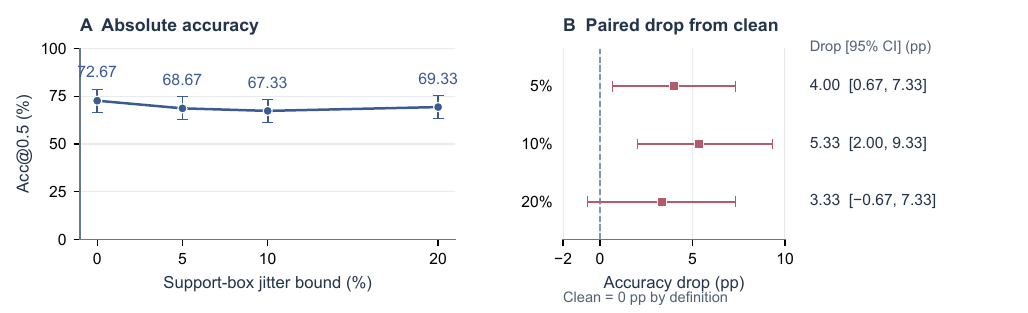}
\caption{Matched support-box sensitivity on the same 150 pairs. The panels give absolute accuracy and clean-minus-perturbed drops with stratified 95\% intervals. All conditions share the checkpoint and renderer.}\label{fig:matched-noise}
\end{figure*}
\endgroup
\subsection{Scope and domain sensitivity}
We apply two post-hoc filters to the canonical predictions: exclude Anime-Style episodes, or retain pairs whose curated strata agree with compatible construction-time threshold assignments (Appendix~\ref{app:historical-audit}). Figure~\ref{fig:scope-sensitivity} combines subset sizes, absolute scores and paired contrasts. The filters overlap; each defines a descriptive evaluation scope. Scope and domain intervals use 5,000 paired target-base-pair-cluster resamples (seed 20261009), with episode-weighted aggregation and no multiplicity adjustment. Each resampled base pair contributes all of its associated episodes together, retaining the dependence between different interaction histories built from the same visual pair.

\begin{figure}[pos=htbp]
\centering\includegraphics[width=\linewidth]{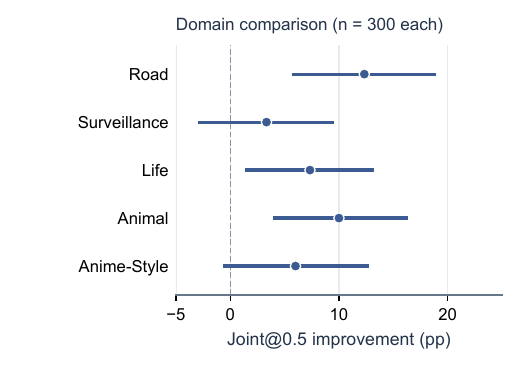}
\caption{Within-domain localization gains. Points show TT-VG minus Seq-free Joint@0.5 with paired base-pair-cluster 95\% intervals; each domain contains 300 episodes. Variation across domains characterizes the curated evaluation sample.}\label{fig:domain-sensitivity}
\end{figure}

Figure~\ref{fig:domain-sensitivity} separates the five visual domains. Within-domain uncertainty is wider than for the full set; the Surveillance and Anime-Style intervals include zero. Residual dependence between source scenes may remain.

\begingroup
\makeatletter
\def\@dbflt#1{\@ifnextchar[{\@xdblfloat{#1}}{\@xdblfloat{#1}[t]}}
\makeatother
\begin{figure*}[pos=t]
\centering\includegraphics[width=\textwidth]{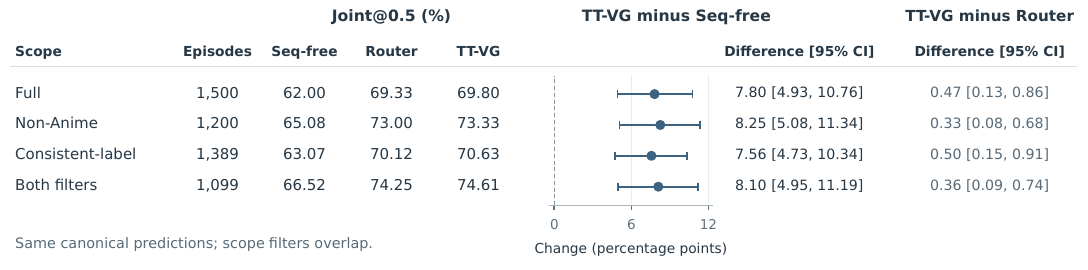}
\caption{Post-hoc scope sensitivity on fixed canonical predictions. Scores are percentages; points show TT-VG minus Seq-free with paired 95\% intervals, and the right column gives the Router contrast. The TT-VG--Seq-free difference remains positive across the overlapping scopes.}\label{fig:scope-sensitivity}
\end{figure*}
\begin{figure*}[pos=t]
\centering\includegraphics[width=\textwidth]{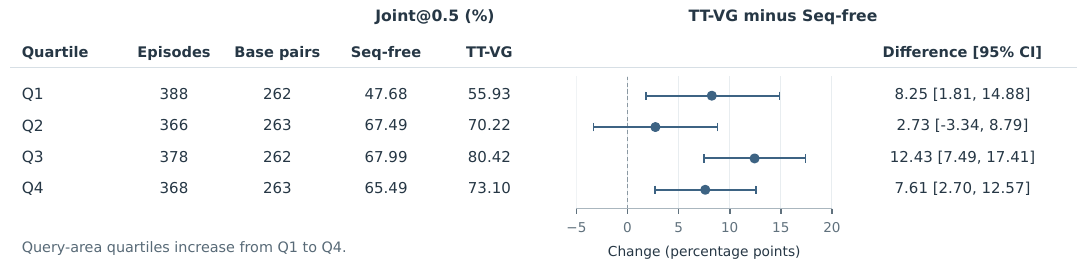}
\caption{Localization scores and gains across query-area quartiles. Points show TT-VG minus Seq-free Joint@0.5 with paired 95\% intervals. The quartiles describe geometry separately from the curated strata.}\label{fig:geometric-scale}
\end{figure*}
\endgroup

\subsection{Query-area sensitivity}
Query-box area provides a geometric descriptor complementary to the curated strata. Quartile cutpoints are computed over the 1,050 base pairs using query-box area as a proportion of query-image area. The thresholds are 1.43\%, 3.93\% and 9.79\%. Boundary ties are assigned to the upper bin, and all episodes derived from the same base pair retain the same bin. Figure~\ref{fig:geometric-scale} reports scores, denominators and paired differences. Intervals use 5,000 paired base-pair-cluster resamples (seed 20261012), with episode-weighted aggregation and no multiplicity adjustment. The second-quartile interval includes zero. These post-hoc bins preserve the primary strata; occlusion and appearance ambiguity are separate attributes.

\section{Threshold Sensitivity}\label{app:reading-diagnostics}
The threshold analysis reuses canonical predictions and varies the IoU criterion. Exploratory, unadjusted intervals use 5,000 paired base-pair-cluster resamples (seed 20261012). Small differences from the separately seeded primary intervals reflect Monte Carlo resampling.
\begin{table}[pos=!htbp]\caption{Selected IoU-threshold sensitivity. Differences are TT-VG minus Seq-free joint accuracy, in percentage points.}\centering\footnotesize\paperTableStyle\footnotesize\begin{tabular*}{\linewidth}{@{\extracolsep{\fill}}rrr@{}}\toprule
IoU threshold & Difference & Paired 95\% CI\\\midrule
0.30 & 5.53 & [2.81, 8.34]\\
0.50 & 7.80 & [4.87, 10.68]\\
0.70 & 9.93 & [6.91, 13.07]\\
0.90 & 8.00 & [5.21, 10.96]\\
\bottomrule\end{tabular*}\end{table}

\section{Complete Dependency-Group Results}\label{app:dependency-groups}
Table~\ref{tab:all-dependency-counts} reports all three configurations across the four dependencies partitioning 1,500 canonical episodes.

\par

\begin{table}[pos=!htbp]
\caption{Joint@0.5 success counts for all four dependencies. The three configurations share the same $n$ within each row.}\label{tab:all-dependency-counts}
\centering\footnotesize\paperTableStyle
\begin{tabular*}{\linewidth}{@{\extracolsep{\fill}}lrrrr@{}}\toprule
Dependency & $n$ & Seq-free & Router & TT-VG \\
\midrule
Explicit-Name & 600 & 380 & 406 & 406 \\
Active-Target & 200 & 142 & 139 & 139 \\
Counter-Recency & 609 & 357 & 436 & 438 \\
Rollback & 91 & 51 & 59 & 64 \\
\bottomrule\end{tabular*}
\par\smallskip\parbox{\linewidth}{\footnotesize Router denotes Pred-Name Router + VEGA.}
\end{table}

With VEGA fixed, TSTT adds two Counter-Recency successes (+0.33 points) and five Rollback successes (+5.49 points), leaving other groups unchanged. Thus, five of seven additional successes involve state restoration.
\section*{Declaration of generative AI and AI-assisted technologies in the manuscript preparation process}
GPT (OpenAI) assisted with code review, result auditing, manuscript language polishing, and plotting-code preparation. The authors retain responsibility for verifying the outputs and for the manuscript content. The Methods detail research-image and controlled wording generation.
\par
\raggedbottom
\nobalance
\begingroup
\interlinepenalty=10000
\bibliographystyle{scvib-num-link}
\bibliography{references}
\par
\endgroup
\end{document}